\documentclass[conference]{IEEEtran}

\usepackage{xargs}
\usepackage{hyperref}

\usepackage{amssymb}
\usepackage{amsmath}

\usepackage{tikz}
\usetikzlibrary{arrows}
\usetikzlibrary{positioning,fit,calc}

\usepackage{graphicx}
\usepackage{subcaption}
\graphicspath{ {figures/} }

\usepackage{xcolor}

\newcommandx{\spider}{SPIDER}
\newcommandx{\fractal}{FRACTAL}
\newcommandx{\selene}{SELENE}

\newcommand{\nats}{\mathbb{N}}
\newcommand{\reals}{\mathbb{R}}

\def\BibTeX{{\rm B\kern-.05em{\sc i\kern-.025em b}\kern-.08em
    T\kern-.1667em\lower.7ex\hbox{E}\kern-.125emX}}

\begin{document}
    \title{Reinforcement learning for safety-critical control of an automated vehicle}

	\author{
		\IEEEauthorblockN{1\textsuperscript{st} Florian Thaler}
		\IEEEauthorblockA{\textit{Virtual Vehicle Research GmbH} \\
		Graz, Austria \\
		florian.thaler@v2c2.at}
		\and{}
		\IEEEauthorblockN{2\textsuperscript{nd} Franz Rammerstorfer}
		\IEEEauthorblockA{\textit{Virtual Vehicle Research GmbH} \\
		Graz, Austria \\
		franz.rammerstorfer@v2c2.at}
		\and
		\IEEEauthorblockN{3\textsuperscript{nd} Jon Ander Gomez}
		\IEEEauthorblockA{\textit{Solver Intelligent Analytics} \\
		Valencia, Spain \\
		jon@iasolver.com}
		\and
		\IEEEauthorblockN{4\textsuperscript{rd} Raul Garcia Crespo}
		\IEEEauthorblockA{\textit{Solver Intelligent Analytics} \\
		Valencia, Spain \\
		rgarcia@iasolver.com}
		\and
		\IEEEauthorblockN{5\textsuperscript{th} Leticia Pasqual}
		\IEEEauthorblockA{\textit{Solver Intelligent Analytics} \\
		Valencia, Spain \\
		lpascual@iasolver.com}
		\and
		\IEEEauthorblockN{6\textsuperscript{th} Markus Postl}
		\IEEEauthorblockA{\textit{Virtual Vehicle Research GmbH} \\
		Graz, Austria \\
		markus.postl@v2c2.at}
	}
    \maketitle

    \begin{abstract}
		We present our approach for the development, validation and 
		deployment of a data-driven decision-making function for the 
		automated control of a vehicle. The decision-making function, 
		based on an artificial neural network is trained to steer the 
		mobile robot {\spider} towards a predefined, static path to a 
		target point while avoiding collisions with obstacles along the 
		path. The training is conducted by means of proximal policy 
		optimisation (PPO), a state of the art algorithm from the field 
		of reinforcement learning. 
		\newline
		The resulting controller is validated using KPIs quantifying
		its capability to follow a given path and its reactivity on
		perceived obstacles along the path. The corresponding tests are
		carried out in the training environment. 
		Additionally, the tests shall be performed as well in the 
		robotics situation Gazebo and in real world scenarios. For the 
		latter the controller is deployed on a FPGA-based development
		platform, the {\fractal} platform, and integrated into 
		the {\spider} software stack.
    \end{abstract}

	\begin{IEEEkeywords}
		Reinforcement learning, Decision-making, Path following, 
		Path tracking, Reactive path tracking, Collision avoidance, 
		Automated driving, Validation, EDDL
	\end{IEEEkeywords}

    \section{Introduction}
		In this work we aim to showcase the implementation, validation and 
deployment of a machine learning (ML) application in a safety critical
system. For this purpose a data-driven decision-making function for the
automated control of the mobile robot {\spider} is developed. It extends
the capabilities of a Stanley-based path tracking 
controller\footnote{For a comprehensive
description and discussion of how a Stanley controller works we refer to
\cite{hoffmann2007autonomous} and \cite{dominguez2016comparison}.} which
is already integrated in the {\spider} software stack\footnote{The 
software stack of the {\spider} is entirely based on ROS 2.} to a 
reactive path tracking controller. Thus, if an obstacle along the path 
is perceived by the robot, an evasion maneuver to avoid a 
collision is initiated.
\newline
\newline
The function is composed of several function blocks (see section 
\ref{subsec:design}). Its decision-making block, i.e.\@ the unit which 
is providing the controls to be applied to the vehicle, is based on 
an artificial neural network (ANN). For the training procedure of 
this ANN, a state of the art algorithm from the field of 
reinforcement learning (RL) is used - see section \ref{subsec:training}.

The focal point of this work is on investigating the 
preservation of safety relevant driving functions\footnote{This relates
in particular to the collision avoidance function, which triggers an 
emergency brake if a collision is imminent.} while executing the 
abovementioned decision-making function. Hence, a framework supporting 
on one hand the execution of computationally intense vehicle functions,
and on the other hand allowing its safe execution has to be provided. 
This is where the {\spider} and the {\fractal} project come into play.
 
The {\spider}\footnote{\url{https://www.v2c2.at/spider/}} is a 
mobile HiL platform developed at the Virtual Vehicle Research GmbH. 
It is designed for the testing of autonomous driving
functions in real-world conditions, e.g.\@ on proving grounds, in an 
automated and reproducible manner. The integrated safety concept
ensures the safety of test drives. The {\fractal} 
platform\footnote{See \url{https://fractal-project.eu/} and \cite{9217800}}
a FPGA-based development platform, and various components developed in 
the {\fractal} project enhance the already existing safety concept. 
This includes monitoring units and a diverse redundancy library. 
In addition, the integrated hardware accelerators make the platform 
suitable for the execution of functions with high computing effort. 
\newline
Due to the open system design of the {\spider}, the {\fractal} platform
can be integrated into the {\spider} system. To incorporate the 
decision-making function into the {\spider} software stack and to run 
it on the {\fractal} platform, it is integrated into an appropriate
ROS 2 node. For the deployment of the ANN the open-source deep learning 
library EDDL\footnote{See \url{https://github.com/deephealthproject/eddl/}} 
is used.

\subsection{Related work}
	The scientific literature knows several non data-driven methods for 
	the solution of the path tracking and obstacle avoidance problem. 
	Some popular and well-known path tracking controllers are 
	the “Pure pursuit controller”, the “Carrot chasing controller”, or 
	the “Stanley controller” - for details we refer to
	\cite{gutierrez2020waypoint}, \cite{perez2019aerial}, 
	\cite{samuel2016review} or \cite{hoffmann2007autonomous}. Approaches
	for the design of obstacle avoidance controllers, such as the 
	artificial potential field method, can be found in 
	\cite{rostami2019obstacle}, \cite{wiig20183d}, \cite{leca2019sensor}.
	\newline
	However, we decided to follow a ML approach to tackle 
	the reactive path tracking task. The main reason for this decision
	is that it seemed us to be difficult to appropriately tune and
	coordinate a combined controller consisting of a path tracking 
	and a collision avoidance component. In addition, a slim and 
	efficient ML solution promises a low computational effort at 
	runtime. Quoting \cite{kober2013reinforcement}, RL offers to 
	robotics a framework and a set of tools for the design of 
	sophisticated and hard-to-engineer behaviors. According to that, RL
	approaches are well suited to the problem. The application of RL 
	methods for the automated control of vehicles 
	is not new - the topic was already adressed by a variety of 
	researchers. We refer in this regard to \cite{folkers2019controlling}, 
	\cite{kiran2021deep}, \cite{ultsch2020reinforcement}, 
	\cite{alomari2021path}, \cite{cheng2022path} and as well 
	\cite{meyer2020taming}.
	For the sake of completeness, we point out that there are also 
	non machine learning based methods for solving the reactive path 
	tracking problem. See for example \cite{vougioukas2007reactive}, 
	\cite{hassani2018robot}.
	
\subsection{Structure of the paper}

	This paper is structured as follows. 
	In Section \ref{sec:preliminaries} the main building blocks of this
	work are presented. Section \ref{sec:problemFormulation} gives the 
	formulation of the problem. In Section \ref{sec:decisionMaking} 
	the structure 	and functioning of the decision-making function is 
	presented. Finally, in the remaining sections the results obtained are 
	presented and discussed.

	\section{Preliminaries}\label{sec:preliminaries}
		\subsection{\spider}
\begin{figure}[ht]
		\centering
		\includegraphics[width = .45\textwidth]{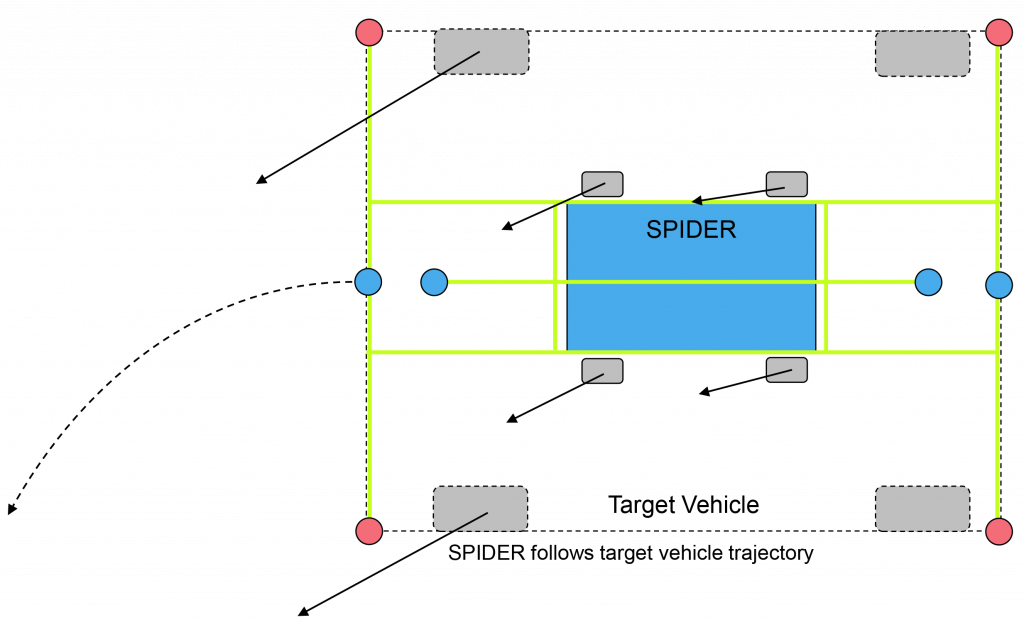}
		\caption{SPIDER following the trajectory of a target vehicle. 
			The red dots indicate SPIDER sensors ensuring safety, where 
			the blue dots correspond to sensors of the target vehicle. 
			The green bars represent rods on which the 
			sensors are mounted.}\label{fig:spider_target}
	\end{figure}

	The {\spider} (Smart Physical Demonstration and Evaluation Robot) is 
	an autonomous robot prototype developed at Virtual Vehicle
	Research GmbH. It is a mobile HiL platform designed for the 
	development and testing of autonomous driving functions. It allows
	reproducible testing of perception systems, vehicle software and 
	control algorithms under real world conditions. Four individually
	controllable wheels enable almost omni-directional movement, 
	enabling the {\spider} to precisely mimic the movements of target 
	vehicles - see Figure \ref{fig:spider_target}.
	
\begin{figure}[ht]
		\centering
		\includegraphics[width = .45\textwidth]{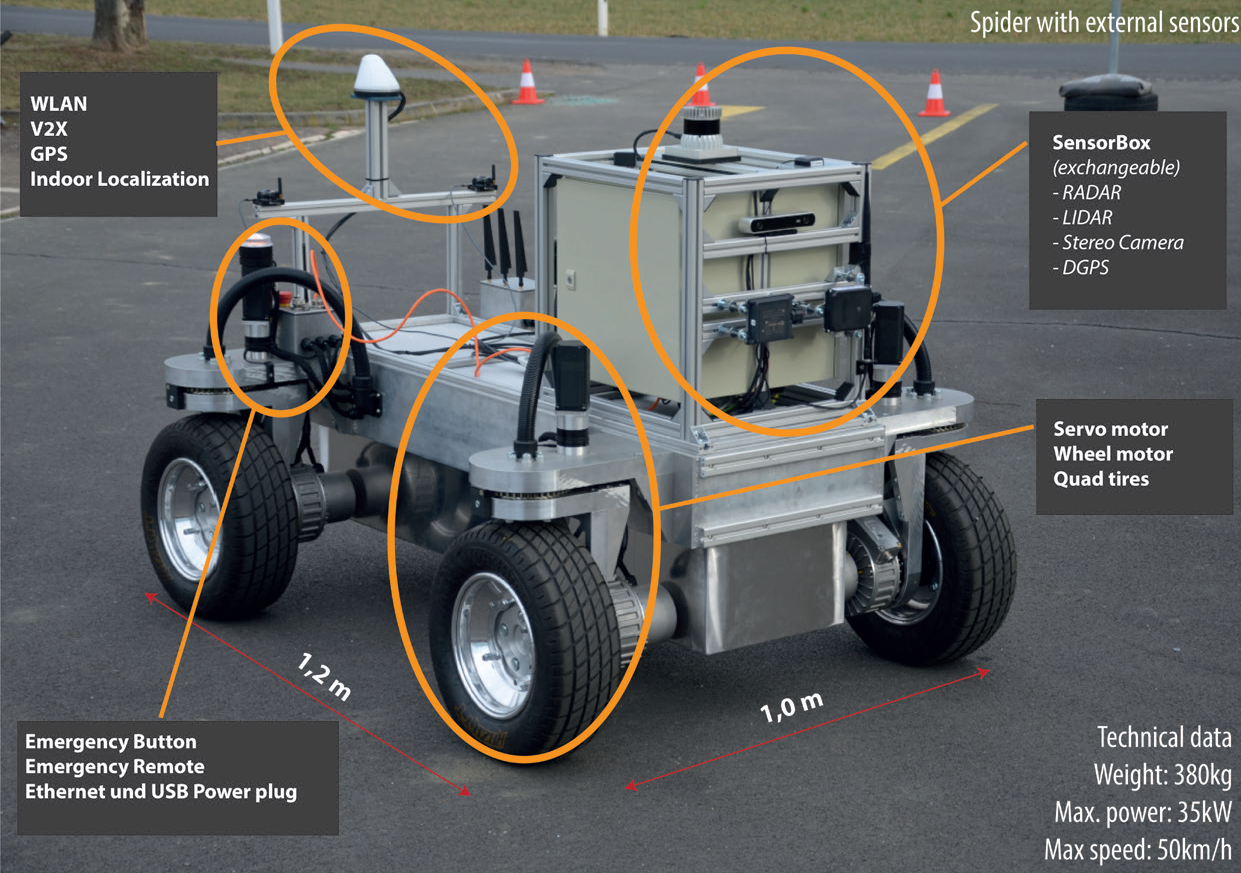}
		\caption{Smart Physical Demonstration and Evaluation Robot 
			({\spider})}\label{fig:spider_with_facts}
	\end{figure}

	Due to its adaptable mounting rod system - see Figure
	\ref{fig:spider_with_facts} - positions of sensors can easily be 
	adapted to the target system.
	
	From a system perspective, the architecture of the {\spider} can be 
	divided into three blocks, as shown in Figure 
	\ref{fig:system_architecture}. The decision-making function to be
	developed is located in the HLCU. Using the data provided by the 
	sensor block, it determines control variables which are passed
	on to the LLCU in the form of a target linear speed and target 
	angular velocity. The LLCU performs safety 
	checks on these signals and forwards them to the corresponding 
	hardware components.
	

\definecolor{lightGray}{rgb}{0.83, 0.83, 0.83}
\definecolor{lightGreen}{rgb}{0.65, 0.89, 0.62}

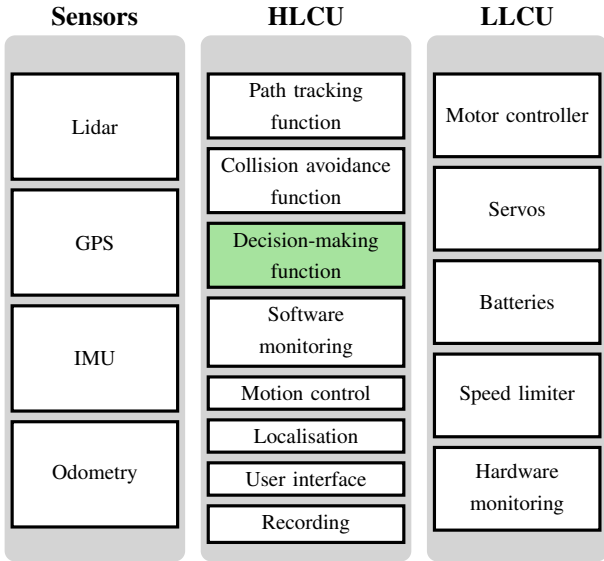
\begin{figure}
	\centering
	\begin{tikzpicture}[auto, >=latex']

        \node (block_1) [draw = none, fill = lightGray, 
			minimum width = 24mm, minimum height = 70mm, 
			rounded corners]{};
        \node (block_2) [right = 16mm of block_1.east, 
			anchor = center, draw = none, 
			fill = lightGray, minimum width = 28mm, 
			minimum height = 70mm, rounded corners]{};
		\node (block_3) [right = 14mm of block_2.east, 
			anchor = center, draw = none, 
			fill = lightGray, minimum width = 24mm, 
			minimum height = 70mm, rounded corners]{};
		
        \node (block_1_label)[above = 0mm of block_1] {\textbf{Sensors}};
        \node (block_2_label)[above = 0mm of block_2] {\textbf{HLCU}};
        \node (block_3_label)[above = 0mm of block_3] {\textbf{LLCU}};
        
        
        \node (node_1_block_1) [very thick, draw = black, 
			below = 5mm of block_1.north, 
			anchor = north, fill = white, minimum width = 22mm, 
			minimum height = 14mm]
			{\footnotesize Lidar};
		\node (node_2_block_1) [very thick, draw = black,
			below = 1mm of node_1_block_1, 
			fill = white, minimum width = 22mm, 
			minimum height = 14mm]
			{\footnotesize GPS};
		\node (node_3_block_1) [very thick, draw = black,
			below = 1mm of node_2_block_1, 
			fill = white, minimum width = 22mm, 
			minimum height = 14mm]
			{\footnotesize IMU};
		\node (node_4_block_1) [very thick, draw = black,
			below = 1mm of node_3_block_1, 
			fill = white, minimum width = 22mm, 
			minimum height = 14mm]
			{\footnotesize Odometry};
			
		\node (node_1_block_2) [very thick, draw = black, 
			below = 5mm of block_2.north, align = center, 
			anchor = north, fill = white, minimum width = 26mm]
			{\footnotesize Path tracking\\\footnotesize function};
		\node (node_2_block_2) [very thick, draw = black, 
			below = 1mm of node_1_block_2, align = center, 
			fill = white, minimum width = 26mm]
			{\footnotesize Collision avoidance\\\footnotesize function};
        \node (node_3_block_2) [very thick, draw = black, 
			below = 1mm of node_2_block_2, align = center, 
			fill = lightGreen, minimum width = 26mm]
			{\footnotesize Decision-making\\\footnotesize function};
		\node (node_4_block_2) [very thick, draw = black, 
			below = 1mm of node_3_block_2, align = center, 
			fill = white, minimum width = 26mm]
			{\footnotesize Software\\\footnotesize monitoring};
		\node (node_5_block_2) [very thick, draw = black, 
			below = 1mm of node_4_block_2, 
			fill = white, minimum width = 26mm]
			{\footnotesize Motion control};
		\node (node_6_block_2) [very thick, draw = black, 
			below = 1mm of node_5_block_2, 
			fill = white, minimum width = 26mm]
			{\footnotesize Localisation};
		\node (node_7_block_2) [very thick, draw = black, 
			below = 1mm of node_6_block_2, 
			fill = white, minimum width = 26mm]
			{\footnotesize User interface};
		\node (node_8_block_2) [very thick, draw = black, 
			below = 1mm of node_7_block_2, 
			fill = white, minimum width = 26mm]
			{\footnotesize Recording};
		
        \node (node_1_block_3) [very thick, draw = black, 
			below = 5mm of block_3.north, 
			anchor = north, fill = white, minimum width = 22mm,
			minimum height = 11mm]
			{\footnotesize Motor controller};
		\node (node_2_block_3) [very thick, draw = black, 
			below = 1mm of node_1_block_3, 
			fill = white, minimum width = 22mm, 
			minimum height = 11mm]
			{\footnotesize Servos};
		\node (node_3_block_3) [very thick, draw = black, 
			below = 1mm of node_2_block_3, 
			fill = white, minimum width = 22mm, 
			minimum height = 11mm]
			{\footnotesize Batteries};
		\node (node_4_block_3) [very thick, draw = black, 
			below = 1mm of node_3_block_3, 
			fill = white, minimum width = 22mm, 
			minimum height = 11mm]
			{\footnotesize Speed limiter};
		\node (node_5_block_3) [very thick, draw = black, 
			below = 1mm of node_4_block_3, align = center,  
			fill = white, minimum width = 22mm, 
			minimum height = 11mm]
			{\footnotesize Hardware\\\footnotesize monitoring};
	\end{tikzpicture}
	\caption{System architecture of the {\spider}.}
	\label{fig:system_architecture}
\end{figure}

\subsection{The {\fractal} platform}
\begin{figure}[ht]
		\centering
		\includegraphics[width = 0.45\textwidth]{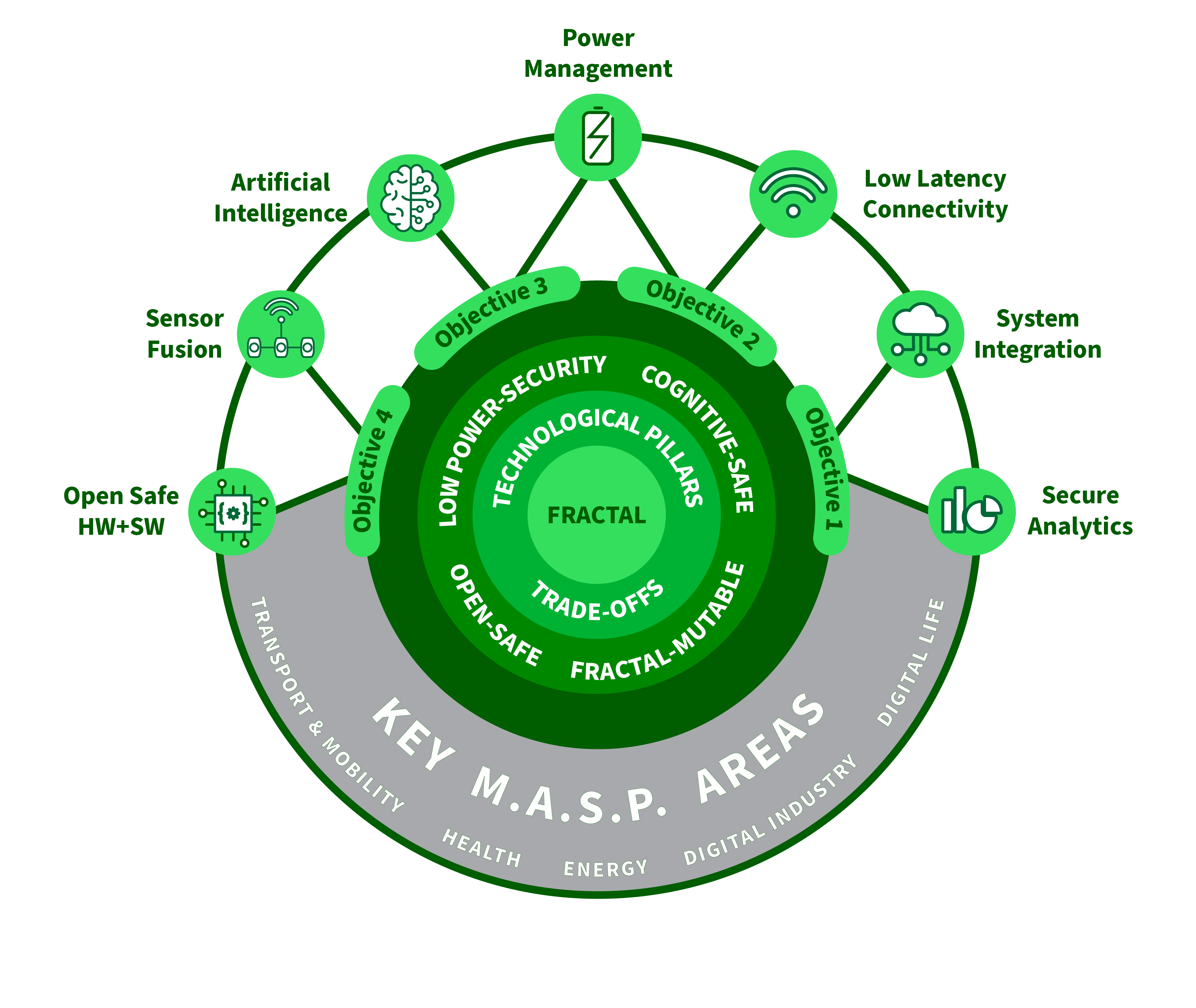}
		\caption{{\fractal} project technology pillars and objectives
			under the Multi-Annual Strategic Plan ("M.A.S.P.")}
			\label{fig:fractal_technology_pillars}
	\end{figure}

	The {\fractal} platform \cite{9217800} is a new approach to reliable
	edge computing. It provides an Open-Safe-Reliable platform to build
	congnitive edge nodes while guaranteeing extra-functional properties
	like dependability, or security, as visualized in Figure 
	\ref{fig:fractal_technology_pillars}. {\fractal} nodes can be deployed
	to various hardware architectures. The {\spider} use-case is deployed
	on a FPGA using the open-source {\selene} hardware and software 
	platform \cite{selene2020}. {\selene} is a heterogeneous multicore
	processor platform based on the open RISC-V Instruction Set 
	Architecture (ISA). The software stack is build on GNU/Linux. 
	The {\selene} platform is extended by various components from {\fractal}
	to ensure safety properties and allow the execution of computational 
	extensive machine learning functions.
	
	The developments of {\selene} and {\fractal} provide the baseline for 
	the {\spider} to move from a non-safe industrial PC setup to an 
	open source based, safe platform with smaller form-factor, and 
	lower power comsumption. To add extra properties in context of 
	safety and hardware acceleration, the {\spider} includes {\fractal}
	components on hardware and software level. This includes congestion 
	detection at the memory controller, register file randomization, 
	a redundant acceleration scheme, diverse redundancy of cores, 
	and statistics units \cite{safeX2022}. 
	
\subsection{Reinforcement learning}
	Reinforcement learning (RL) refers to a subarea of machine learning.
	The learning principle of methods belonging to this area is based 
	on learning through interaction. Through repeated interaction with 
	its environment, the learning system learns which actions are 
	beneficial in terms of problem solving and which are detrimental
	in this respect. This is done by means of a numerical reward 
	function tailored to the specific use case. By means of suitable 
	optimization methods the system is encouraged to derive a 
	control strategy, which, given a certain observation, selects the 
	action that promises the maximum reward.
	\newline
	A more detailed description of the RL paradigm can be found for
	example in \cite{kober2013reinforcement},
	\cite{sutton2018reinforcement}.
	
\subsection{EDDL/LEDEL}

	The European Distributed Deep Learning Libray (EDDL) is a 
	general-purpose deep learning library initially developed as part 
	of the DeepHealth Toolkit \cite{dh:toolkit} to cover deep learning 
	needs in healthcare use cases within the DeepHealth project 
	\url{https://deephealth-project.eu/}. The EDDL is a free and 
	open-source software available on a GitHub repository 
	\url{https://github.com/deephealthproject/eddl/}.

	EDDL provides hardware-agnostic tensor operations to facilitate 
	the development of hardware-accelerated deep learning 
	functionalities and the implementation of the necessary tensor 
	operators, activation functions, regularization functions, 
	optimization methods, as well as all layer types 
	(dense, convolutional and recurrent) to implement 
	state-of-the-art neural network topologies.
	Given the requirement for fast computation of matrix operations and 
	mathematical functions, the EDDL is being coded in C++. GPU specific 
	implementations are based on the NVIDIA CUDA language extensions 
	for C++. A Python API is also available in the same GitHub 
	repository and known as pyEDDL.
	\newline
	In order to be compatible with existing developments and other 
	deep learning toolkits, the EDDL uses ONNX \cite{onnx}, the 
	standard format for neural network interchange, to import and 
	export neural networks including both weights and topology.
	\newline
	In the Fractal project, the EDDL is being adapted to be 
	executed on embedded and safety-critical systems. Usually, 
	these systems are equipped with low resources, i.e., with 
	limited memory and computing power, as it is the case of 
	devices running on the edge.
	
	When adapted to this kind of systems, the EDDL is renamed 
	as Low Energy DEep Learning library (LEDEL).

	Specifically, the EDDL has been ported to run on emulated 
	environments based on the RISC-V CPU. It has been tested to 
	train models and for inferencing. However, the use of the 
	LEDEL in this work is only for inferencing, so that the
	running time is not a critical issue. The models are trained 
	using the EDDL on powerful computers, then the trained models 
	can be imported by the EDDL thanks to ONNX.

	\section{Problem formulation}\label{sec:problemFormulation}
		The decision-making function shall navigate the {\spider} along a 
predefined path\footnote{By a path we mean a list of target coordinates, 
target linear speeds and target headings which shall be reached 
one after another by the robot. Only static paths are considered,
i.e.\@ any path is generated in advance by a path planning module 
and is not changed during execution time. Furthermore, we assume that 
the target speeds do not exceed the achievable maximum speed of the 
vehicle.} from a starting point to a target point while avoiding 
collisions with obstacles.
\newline
Although the SPIDER can be controlled omnidirectionally, in this 
use case we limit ourselves to develop a car-like control strategy. 
As a consequence, reaching the target orientation at each of the 
given waypoints is disregarded.

	\section{Decision-making function}\label{sec:decisionMaking}
		\subsection{Design}\label{subsec:design}
	The entire control unit is designed as depicted in Figure 
	\ref{fig:decMakFlow}. At any time point it takes as input a 
	cost map, the current state	of the vehicle, the control values
	applied in the previous time step and provides the control values 
	$(u_{1}, u_{2})$ to be applied next as output. These values
	are sampled from the finite subset
	\begin{align*}
		U = \{(u_{1}, &u_{2})~:~u_{1} = -0.5 + 1.5i / 11, \\
			&u_{2} = -1 + 2j / 11 ~ 1\leq i, j\leq 11\}.
	\end{align*}
	of the control space $[-1/2, 1]\times[-1, 1]$ according to the 
	probability distribution provided by the decision-making block 
	represented through an ANN. Given the maximal linear 
	acceleration $a_{max}$ and the maximal steering angle 
	$\vartheta_{max}$ of the robot, the terms $u_{1}a_{max}$, 
	$u_{2}\vartheta_{max}$ determine the acceleration and the steering 
	angle respectively which will be applied to the vehicle.
	
	We briefly discuss the main building blocks of the control unit next.
	\tikzstyle{block} = [draw, rectangle, rounded corners,
    minimum height = 3mm, minimum width = 9mm]
\tikzstyle{arrow} = [thick, -> , >=stealth]

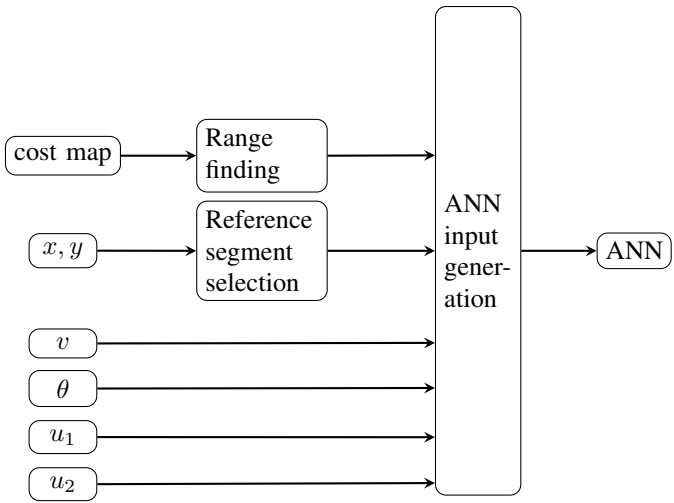
\begin{figure}
	\centering
	\begin{tikzpicture}[auto, >=latex']
		\node [block, anchor = north] (costMap) {cost map};
		\node [block, below = 10mm of costMap.south, anchor = center]
			(pos) {$x, y$};
		\node [block, below = 10mm of pos.south, anchor = center] 
			(speed) {$v$};
		\node [block, below = 4mm of speed.south, anchor = center] 
			(hdng) {$\theta$};
		\node [block, below = 4mm of hdng.south, anchor = center] 
			(accCtrl) {$u_{1}$};
		\node [block, below = 4mm of accCtrl.south, anchor = center] 
			(strngCtrl) {$u_{2}$};
				
		\node[block, right = 10mm of costMap.east, text width = 15mm]
			(rangeFinder) {Range finding};
		\node[block, below = 7.75mm of rangeFinder.south, 
			text width = 15mm, anchor = center]
			(refSecSel) {Reference segment selection};
		
		\node[block, right = 20mm of refSecSel.east, text width = 9mm, 
			minimum height = 65mm, anchor = center]
			(inputGen) {ANN input generation};
		
		\node[block, right = 15mm of inputGen.east, text width = 7.5mm, 
			anchor = center]
			(network) {ANN};

		\draw [arrow] (costMap.east) -- (rangeFinder.west);
		\draw [arrow] (pos.east) -- (refSecSel.west);
		\draw [arrow] (speed.east) -- ($(inputGen.west) + (0, -1.225)$);
		\draw [arrow] (hdng.east) -- ($(inputGen.west) + (0, -1.825)$);
		\draw [arrow] (accCtrl.east) -- ($(inputGen.west) + (0, -2.48)$);
		\draw [arrow] (strngCtrl.east) -- ($(inputGen.west) + (0, -3.09)$);
	
		\draw [arrow] (rangeFinder.east) -- ($(inputGen.west) + (0, 1.27)$);
		\draw [arrow] (refSecSel.east) -- ($(inputGen.west) + (0, 0)$);
	
		\draw [arrow] (inputGen.east) -- (network.west);	
	\end{tikzpicture}
	\caption{Decision-making flow}
	\label{fig:decMakFlow}
\end{figure}

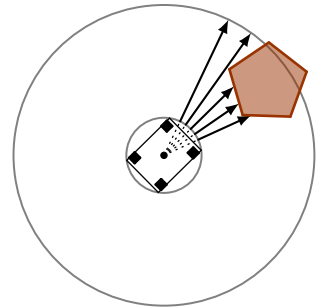
\begin{figure}
	\centering
	\definecolor{darkOrange}{rgb}{0.6,0.2,0}
	\begin{tikzpicture}[line cap=round,line join=round,>=triangle 45,x=1cm,y=1cm]
		
		
		\draw [line width = 0.75pt, color = gray] (2,2) circle (2cm);
		\draw [line width = 0.75pt, color = gray] (2,2) circle (0.5cm);
		
		\draw [line width = 0.5pt] (1.5045265709084834,1.9328726504002063)-- (1.9257096954701793,1.5055498501841997);
		\draw [line width = 0.5pt] (1.9257096954701793,1.5055498501841997)-- (2.495473429091518,2.0671273495997955);
		\draw [line width = 0.5pt] (2.495473429091518,2.0671273495997955)-- (2.074290304529818,2.4944501498157985);
		\draw [line width = 0.5pt] (2.074290304529818,2.4944501498157985)-- (1.5045265709084834,1.9328726504002063);
		
		\draw [-latex, line width = 0.8pt] (2.3561023335133364,2.3509859371347472) -- (2.934868204578819,2.921436233355339);
		\draw [-latex, line width = 0.8pt] (2.411640407279841,2.2838171508110223) -- (3.0016535043674657,2.690618410345189);
		\draw [-latex, line width = 0.8pt] (2.454670995571245,2.2080247239783124) -- (3.162632108377363,2.5319367757990765);
		\draw [-latex, line width = 0.8pt] (2.289744270539504,2.407490193366087) -- (3.158977082158016,3.629960773464348);
		\draw [-latex, line width = 0.8pt] (2.214582474523004,2.451613066271985) -- (2.8583298980920167,3.80645226508794);
		
		\draw [line width = 1pt] (1.5401368042598147,1.967971244113679)-- (1.6103339916867607,1.896750777411015);
		\draw [line width = 1pt] (1.6103339916867607,1.896750777411015)-- (1.6815544583894249,1.9669479648379609);
		\draw [line width = 1pt] (1.6815544583894249,1.9669479648379609)-- (1.6113572709624788,2.038168431540625);
		\draw [line width = 1pt] (1.6113572709624788,2.038168431540625)-- (1.5401368042598147,1.967971244113679);
		
		\fill[line width = 2pt, fill = black,fill opacity = 1] (1.5401368042598147,1.967971244113679) -- (1.6103339916867607,1.896750777411015) -- 			(1.6815544583894249,1.9669479648379609) -- (1.6113572709624788,2.038168431540625) -- cycle;
		
		\draw [line width = 1pt] (1.9674596044758204,2.3891543686753773)-- (2.037656791902766,2.3179339019727134);
		\draw [line width = 1pt] (2.037656791902766,2.3179339019727134)-- (2.108877258605429,2.3881310893996592);
		\draw [line width = 1pt] (2.108877258605429,2.3881310893996592)-- (2.038680071178484,2.4593515561023227);
		\draw [line width = 1pt] (2.038680071178484,2.4593515561023227)-- (1.9674596044758204,2.3891543686753773);
		
		\fill[line width = 2pt, fill = black, fill opacity = 1] (1.9674596044758204,2.3891543686753773) -- (2.037656791902766,2.3179339019727134) -- (2.108877258605429,2.3881310893996592) -- (2.038680071178484,2.4593515561023227) -- cycle;
		
		\draw [line width = 1pt] (1.9613199288215144,1.540648443897676)-- (2.032540395524183,1.6108456313246267);
		\draw [line width = 1pt] (2.032540395524183,1.6108456313246267)-- (1.9623432080972325,1.6820660980272955);
		\draw [line width = 1pt] (1.9623432080972325,1.6820660980272955)-- (1.8911227413945637,1.6118689106003448);
		\draw [line width = 1pt] (1.8911227413945637,1.6118689106003448)-- (1.9613199288215144,1.540648443897676);
		
		\fill[line width = 2pt, fill = black, fill opacity = 1] (1.9613199288215144,1.540648443897676) -- (2.032540395524183,1.6108456313246267) -- (1.9623432080972325,1.6820660980272955) -- (1.8911227413945637,1.6118689106003448) -- cycle;
		
		\draw [line width = 1pt] (2.388642729037514,1.9618315684593681)-- (2.459863195740182,2.0320287558863193);
		\draw [line width = 1pt] (2.459863195740182,2.0320287558863193)-- (2.389666008313231,2.1032492225889876);
		\draw [line width = 1pt] (2.389666008313231,2.1032492225889876)-- (2.3184455416105627,2.0330520351620365);
		\draw [line width = 1pt] (2.3184455416105627,2.0330520351620365)-- (2.388642729037514,1.9618315684593681);
		
		\fill[line width = 2pt, fill = black, fill opacity = 1] (2.388642729037514,1.9618315684593681) -- (2.459863195740182,2.0320287558863193) -- (2.389666008313231,2.1032492225889876) -- (2.3184455416105627,2.0330520351620365) -- cycle;
		
		\draw [line width = 0.5pt, dotted] (2,2)-- (2.214582474523004,2.451613066271985);
		\draw [line width = 0.5pt, dotted] (2,2)-- (2.289744270539504,2.407490193366087);
		\draw [line width = 0.5pt, dotted] (2,2)-- (2.3561023335133364,2.3509859371347472);
		\draw [line width = 0.5pt, dotted] (2,2)-- (2.411640407279841,2.2838171508110223);
		\draw [line width = 0.5pt, dotted] (2,2)-- (2.454670995571245,2.2080247239783124);
				
		\draw [fill = black] (2,2) circle (1.2pt);
		
		\draw [line width = 1pt, color = darkOrange] (3.0464737807620303,2.5357142823881933)-- (3.680683080091598,2.51508958972707);
		\draw [line width = 1pt, color = darkOrange] (3.680683080091598,2.51508958972707)-- (3.8962797799270086,3.1118850960133915);
		\draw [line width = 1pt, color = darkOrange] (3.8962797799270086,3.1118850960133915)-- (3.3953165689580334,3.5013496958926633);
		\draw [line width = 1pt, color = darkOrange] (3.3953165689580334,3.5013496958926633)-- (2.8701075776305123,3.14525654974661);
		\draw [line width = 1pt, color = darkOrange] (2.8701075776305123,3.14525654974661)-- (3.0464737807620303,2.5357142823881933);

		\fill[line width = 2pt, color = darkOrange, fill = darkOrange,fill opacity = 0.5] (3.0464737807620303,2.5357142823881933) -- (3.680683080091598,2.51508958972707) -- (3.8962797799270086,3.1118850960133915) -- (3.3953165689580334,3.5013496958926633) -- (2.8701075776305123,3.14525654974661) -- cycle;	
	\end{tikzpicture}
	\caption{Illustration of the range finding procedure.}
	\label{fig:range_finding}
\end{figure}

	\vspace{.5\baselineskip}
	\textbf{Range finding}
	The range finding block consists of a module which takes as input
	a cost map of a fixed dimension centered around the vehicle and 
	determines the distance from obstacles to the vehicle by means of 
	a ray-casting approach. 
	For this purpose, starting from the center of mass (COM) of the 
	robot, $m$ virtual rays are plotted on the occupancy grid - see 
	Figure \ref{fig:range_finding}. Along each of these arrows, 
	the corresponding cell entry of the occupancy grid is checked 
	at $n$ evenly distributed points, the so-called ray nodes. 
	Based on the number of free cells counted from the inside to the 
	outside, the distance (in meters)
	along a ray from the robot to any obstacles is determined.
	\newline
	It is assumed that the robot is entirely contained within a 
	circular disk of radius $\rho_{1}$ centered at its COM. For the
	determination of the distances we thus only take in consideration
	ray nodes which are not contained in this disk. In addition the
	maximal distance is bounded by $\rho_{2}$. Thus we get distances
	in the interval $[0, \rho_{2} - \rho_{1}]$.
	
	\vspace{.5\baselineskip}
	\textbf{Reference segment selection} 
	To quantify the spatial proximity of the vehicle to the path the 
	cross track error is used. By definition, the cross track error is
	the normal distance from the current position of the vehicle to the 
	target trajectory. In the given context, it is determined as the
	normal distance of the position of the vehicle to the closest line
	segment which is connecting two consecutive waypoints.\footnote{
	By waypoints we understand the target coordinates defined by the 
	path.} For the reference segment selection procedure we followed the 
	apporach described in \cite{ducard2009fault}, Section 9.3.
	
	\vspace{.5\baselineskip}
	\textbf{ANN input generation}	
	The neural network takes $7$ input variables. Let $(x, y)$, $v$, 
	$\theta$ denote the current position, the current linear speed 
	and the current heading of the vehicle respectively. We denote by
	$u_{0}$ $u_{1}$ the control values, which were applied in the 
	previous time step. 
	
	Let $k\in\nats$ be such that the line segment $L_{(x, y)}$ determined by the 
	reference segment selection procedure described above, connects 
	the waypoints $w_{k}, w_{k + 1}$. Let $v_{k + 1}$ denote the 
	target velocity at $w_{k + 1}$ and let $d = (d_{1}, \ldots, d_{m})$ 
	be the tuple of distances computed by 
	the range finding unit given the current position of the vehicle.
	Then the input $x = (x_{1}, \ldots, x_{7})$ to the neural 
	network is defined as follows:
	\begin{itemize}
		\item Clipped cross track error: Let $e_{x}$ denote the 
			signed normal distance of the vehicle's position to 
			$L_{(x, y)}$, i.e.\@ $e_{x}$ defines the current cross track 
			error. Given the clip parameter $\delta > 0$, we define
			\begin{align*}
				x_{1} = 
				\begin{cases}
					-\delta, \textnormal{ if } e_{x} < -\delta\\
					\delta, \textnormal{ if } e_{x} > \delta\\
					e_{x}, \textnormal{ else }
				\end{cases}
			\end{align*}
		\item Linear speed error: The linear speed error is
			defined as $x_{2} = v_{k + 1} - v$. 
		\item Waypoint heading error: We introduce the heading error 
			$x_{3}$ as the cosine of the angle between the vector 
			$v(\theta)$ indicating the driving direction of the vehicle 
			and the vector connecting $w_{k}$ and $w_{k + 1}$. 
		\item Previous control values: Define $x_{4} = u_{0}$ and
			$x_{5} = u_{1}$.
		\item Obstacle heading error: We introduce $x_{6}$ to be the 
			cosine of the angle between $v(\theta)$ and the range finding
			ray sensing the smallest distance to an obstacle.
		\item Smallest obstacle distance: Define $x_{7}$ to be the smallest 
			distance to an object measured by the range finding unit.
	\end{itemize}
	We note that by these definitions the input variables of the neural 
	network are always contained within a fixed range and are thus 
	bounded. According to \cite{sola1997importance} such normalisation 
	can accelerate and stabilise the training process.
	
	\vspace{.5\baselineskip}
	\textbf{Decision-making}
	The ANN representing the decision-making block consists of two 
	hidden dense layers of $64$ neurons each and an output layer of 
	$|U| = 121$ neurons. For the hidden layers $\tanh$ is used as
	activation function, whereas for the output layer the softmax function
	is used.

\subsection{Training}\label{subsec:training}
	
	For the training of the decision-making function we use the proximal 
	policy optimisation (PPO) method presented in 
	\cite{schulman2017proximal}. The training suite is implemented in
	Python and is built on the Python package Stable Baselines - see
	\cite{stable-baselines}. The driving environment which is used 
	for the training procedure is based on the well known kinematic bicycle 
	model\footnote{For a description and an analysis of the model we 
	refer to \cite{kong2015kinematic}.}, which is - according to
	\cite{polack2017kinematic} - a suitable and accurate model for 
	car-like driving manoeuvres at low speeds. 
	\newline
	The RL paradigm requires to choose a reward function adapted and 
	aligned to the problem. In the given context it must be constructed 
	in such a way that actions steering the vehicle with target speed 
	along the target trajectory are rewarded. 
	However, the rewarding approach must also reflect the collision 
	avoidance requirement. Thus, actions which may lead to 
	collisions must be penalised. 
	
	The approach we considered, builds on the work of 
	\cite{ultsch2020reinforcement}, \cite{cheng2022path} and 
	\cite{meyer2020taming}. Its main lines are described next. 
	The reward function $r$ is made up of a path following component 
	$r_{pf}$ and a collision avoidance component $r_{ca}$. Given the 
	input data $x = (x_{1}, \ldots, x_{7})$ to the ANN,
	non-negative parameters $\alpha_{1}, \ldots, \alpha_{4}$, 
	$\beta_{1}, \beta_{2}$, $\lambda$ we define
	\begin{align*}
		&r_{1}(x) = \alpha_{1}\exp(-x_{1}^{2} / 2\beta_{1})\\
		&r_{2}(x) = \alpha_{2}\exp(-x_{2}^{2} / 2\beta_{2})\\
		&r_{3}(x) = \alpha_{3} x_{3}
	\end{align*}
	With this we set
	\begin{align*}
		&r_{pf}(x) = -1 + (1 + r_{2}(x) r_{3}(x))(1 + r_{1}(x))\\
		&r_{ac}(x) = 
		\begin{cases}
			-\alpha_{4} x_{6}, \textnormal{ if } x_{7} \leq \lambda (\rho_{2} - \rho_{1})\\
			0, \textnormal{ else }
		\end{cases}
	\end{align*}
	The definition of $r_{pf}$ implies, that large rewards 
	can be achieved for small values of the cross track error, and 
	small deviations from the target velocity and if the vehicle is 
	approaching in a straight line the upcoming waypoint. Due to the 
	additive constants in the definition of the reward function, 
	driving strategies aligned only partially to the desired policy 
	will not be disregarded. Thus, for example, deviations from the 
	speed profile off the target trajectory result in positive 
	rewards.
	
	The definition of $r_{ac}$ on the other hand, punishes actions which
	steer the vehicle in the direction of the smallest distance to an 
	obstacle. If the vehicle crashes into an obstacle an additional
	penalty $r_{crash}$ is applied. Combining $r_{ac}$, $r_{pf}$ 
	and $r_{crash}$ via 
	\begin{equation*}
		r(x) = r_{ac} + r_{pf} + r_{crash}
	\end{equation*}
	we obtain a reward signal which honours actions that maximise the 
	distance to obstacles and proximity to the path in the absence of 
	obstacles in the vehicle's surrounding.

    \section{Evaluation and performance criteria}
		The validation of the decision-making function is based on KPIs. Two 
classes of KPIs are considered: KPIs for the assessment of the path 
tracking capability and KPIs for the assessment of the collision 
avoidance capability.
\noindent
For the validation we choose a specific path and a distribution of 
obstacles on or near the target trajectory. Let 
$(x_{1}^{(k)}, x_{2}^{(k)}, x_{3}^{(k)}, x_{4}^{(k)}, x_{5}^{(k)}, x_{6}^{(k)}, x_{7}^{(k)})_{0\leq k\leq N}$
denote the set of input terms to the ANN obtained by applying the
decision-making function to the specified scenario in an episode of 
$N + 1$ time steps\footnote{If after $M < N$ time steps the terminal 
position is reached or the vehicle collides with an obstacle, then only
the input terms up to time point $M$ are considered.}.

\subsection{Path tracking}
	For the assessment of the path tracking performance we consider on 
	the one hand the mean $l^{2}$ total tracking error $\kappa_{2}$
	which is defined by
	\begin{equation*}
		\kappa_{2} = \frac{1}{N}\sum_{k = 1}^{N}
			\|(x_{1}^{(k)}, x_{2}^{(k)})\|_{2}^{2}.
	\end{equation*}
	In addition we measure the path tracking capability of the 
	decision-making function by means of the waypoint reach rate $\kappa_{reach}$:
	Consider a tuple $(z_{1}, \ldots, z_{L})$ of points on the target
	trajectory arranged from the starting point towards the
	terminal point. Then $\kappa_{reach}$ is defined to be the quotient
	of the number of points $z_{l}$, $1\leq l\leq L$ which could be 
	approximately reached in successive manner and the total number 
	$L$ of points considered.
	

			
\subsection{Collision avoidance}
	To validate the collision avoidance capabilites of the 
	decision-making function we use the metrics $\kappa_{danger}$ and 
	$\kappa_{dist}$. The latter is defined by means of
	\begin{equation*}
		\kappa_{dist} = \min\{x_{7}^{(k)}~:~0\leq k\leq N\}.
	\end{equation*}
	We emphasize at this point, that $\kappa_{dist}$ is indirectly
	proportional to the so called safety cost function introduced in
	\cite{sisbot2007spatial}. The definition of $\kappa_{danger}$ is
	based on the collision danger introduced in \cite{toussaint2009robot}. 
	Let $\eta_{k}$ be given by
	\begin{align*}
		\eta_{k} = 
		\begin{cases}
			1, \textnormal{ if } x_{7}^{(k)} \leq (\rho_{2} - \rho_{1}) / 2\\
			0, \textnormal{ else }
		\end{cases}
	\end{align*}
	and define 
	\begin{equation*}
		\kappa_{danger} = \frac{1}{N}\sum_{k = 1}^{N}\eta_{k}.
	\end{equation*}

    \section{Results}
\begin{figure}[ht!]
	\centering
	\begin{subfigure}[b]{0.4\textwidth}
		\centering
		\includegraphics[width = 1.1\linewidth]{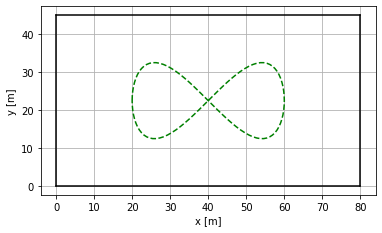}
	\end{subfigure}
	\begin{subfigure}[b]{0.4\textwidth}
		\centering
		\includegraphics[width = 1.1\linewidth]{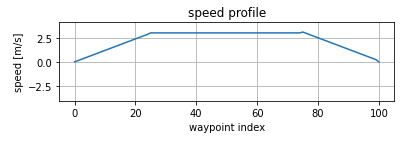}
	\end{subfigure}
	\caption{The plot shows the target trajectory given by the 
		parametric curve $\gamma = (\gamma_{1}, \gamma_{2}):[-\pi, \pi]\to\reals^{2}$,
		where $\gamma_{1}(t) = 40 + 20\cos(t)$, 
		$\gamma_{2}(t) = 22.5 + 20\sin(t)\cos(t)$, and the target
		speed profile.}\label{fig:path}
\end{figure}

		For the evaluation of the decision-making function two scenarios are
examined - a scenario containing obstacles and an obstacle-free 
scenario. In both cases the path depicted in Figure \ref{fig:path} is 
considered.

In the driving simulation, the parameter values 
\begin{equation*}
	a_{max} = 5, \vartheta_{max} = \pi/6, \rho_{1} = 1, \rho_{2} = 5, m = 15, n = 17.
\end{equation*}
were used. In the reference segment selection procedure a lookahead distance 
of $3$ meters was used to obtain the current reference segment and the
corresponding cross track error. For details we refer again to 
\cite{ducard2009fault}. Regarding the reward function we used the 
following values
\begin{table}[h]
	\centering{}
	\begin{tabular}{c|c|c|c|c|c|c|c}
		$\alpha_{1}$ & $\alpha_{2}$ & $\alpha_{3}$ & $\alpha_{4}$ &
		$\beta_{1}$ & $\beta_{2}$ & $\lambda$ & $r_{crash}$\\
		\hline
		$1$ & $1$ & $1$ & $1.5$ & $0.25$ & $0.25$ & $0.75$ & $-250$
	\end{tabular}
\end{table}
The PPO from the Stable Baselines package (version 2.10.0) is applied 
using the default parameter settings.

\subsection{Pure path following}

	\begin{figure}[ht!]
		\centering{}
		\includegraphics[width = 1.0\linewidth]{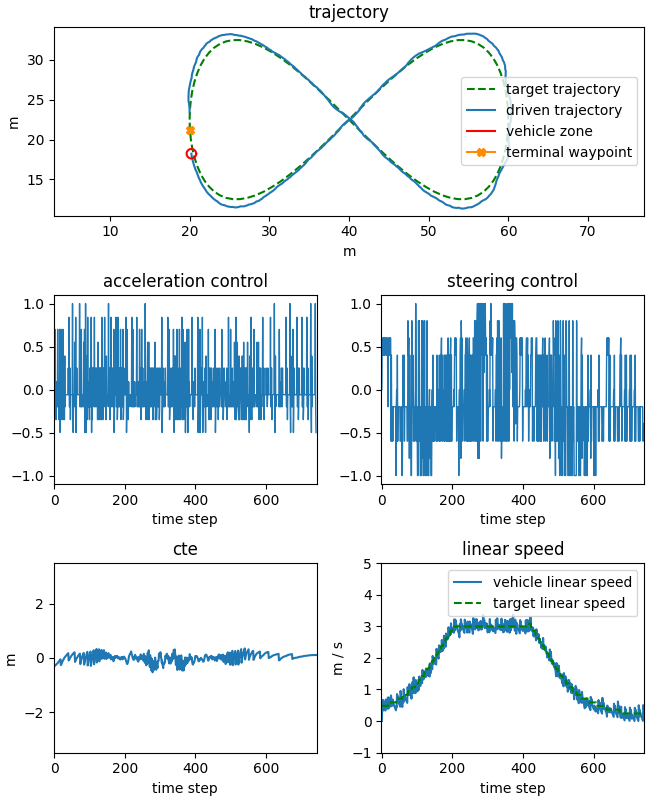}
		\caption{The plot shows the trajectory driven by the robot
			in the Python training simulation in an obstacle-free scenario.
			The plot shows moreover the controls applied over time, 
			the cross track errors at each time point and
			the linear speed profile of the vehicle.}
			\label{fig:results_no_obstacle}
	\end{figure}
	Figure \ref{fig:results_no_obstacle} shows path following
	performance of the decision-making function in an obstacle-free
	environment. We observe that the driven trajectory almost matches
	the target trajectories. The deviations in the curved parts of the
	target trajectory can be attributed to the waypoint selection 
	procedure and the controller's effort to minimise the cross track
	error. Moreover we note, that the target velocity profile is reached
	very precisely. Summing up, we expect a small value of $\kappa_{2}$ 
	and a value close to $1$ of $\kappa_{reach}$. Considering $50$ randomly
	generated points on the target trajectory and a positional tolerance
	of $1$ meters, we obtain
	\begin{equation*}
		\kappa_{2} = 0.04, ~~~~~ \kappa_{reach} = 0.94.
	\end{equation*}
	
\subsection{Reactive path following}
	\begin{figure}[ht!]
		\centering
		\includegraphics[width = 1.0\linewidth]{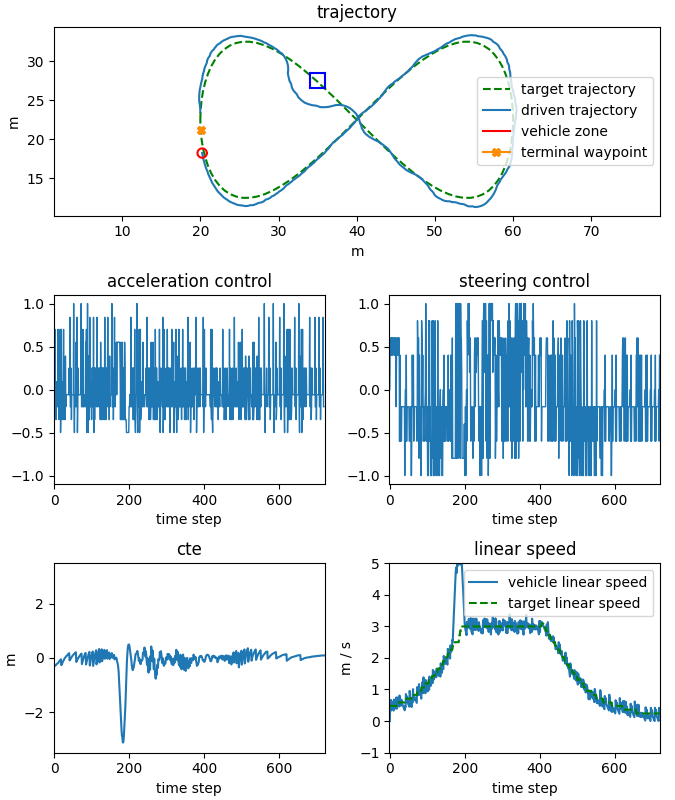}
		\caption{The figure shows the target trajectory and as well 
			the trajectory driven by the vehicle.}
			\label{fig:results_one_obstacle}
	\end{figure}
	To illustrate the ability of the decision-making function to 
	detect and evade an obstacle along the target trajectory, we 
	consider a scenario with one obstacle placed on the target trajectory.
	The resulting trajectory of the vehicle is given in Figure 
	\ref{fig:results_one_obstacle}. In regard of the KPIs we obtain the
	following values
	\begin{align*}
		&\kappa_{2} = 0.36, ~~~~~ \kappa_{reach} = 0.14\\
		&\kappa_{dist} = 0.75, ~~~~~ \kappa_{danger} = 0.02
	\end{align*}
	We point out that, caused by the evasion maneuver, the value of 
	$\kappa_{2}$ increases and the waypoint reach rate $\kappa_{danger}$ 
	drops rather strongly. The latter is a consequence of the necessary 
	wide swerving to avoid a collision.

    \section{Discussion and outlook}
		\subsection{Discussion}
	We were able to define and train by means of a state of the art RL 
	algorithm a decision-making function solving the reactive path tracking
	problem as introduced in Section \ref{sec:problemFormulation}. For the 
	performance and safety assessment of the controller KPIs were 
	considered. According to these KPIs decent results could be obtained. This
	is confirmed by the plots depicted in Figure \ref{fig:results_no_obstacle}
	and Figure \ref{fig:results_one_obstacle}. Even though the target 
	trajectory could be tracked in an sufficient
	manner, by means of adjustments of the reward function a smoother
	driving behaviour could be obtained.
	
	To obtain a more valid and more reliable statement in regard of the 
	safety performance of the decision-making function, further KPIs 
	may be studied. Additionally, unit tests for the examination of the 
	driving behaviours in selected (critical) scenarios could be 
	considered. In order to complete picture the above validation and safety
	assessment procedure has to be applied to a larger set of different 
	scenarios. Only then possible weaknesses of the approach can be 
	identified. Based on these results, conclusions can be drawn about 
	the quality and the completeness of the set of scenarios considered
	in the training process. In order to achieve a balanced and robust 
	result, it is important to use samples from an uniform
	distribution over the whole input space of the ANN. This can be
	achieved by considering a wide variety of training scenarios.
	
\subsection{Outlook}
	At the time of publication of this paper, the integration of the 
	decision-making function into the {\spider} software stack had not 
	yet been completed. Results from the tests carried out in Gazebo
	and in real-world scenarios could therefore not be considered. The
	publication will be supplemented in this respect during the 
	remainder of the {\fractal} project.

    \section{Acknowledgments}
		This project has received funding from the ECSEL Joint Undertaking (JU) 
under grant agreement No 877056. The JU receives support from the
European Union’s Horizon 2020 research and innovation programme and
Spain, Italy, Austria, Germany, Finland, Switzerland. In Austria the
project was also funded by the program "IKT der Zukunft" of the 
Austrian Federal Ministry for Climate Action (BMK).
The publication was written at Virtual Vehicle Research GmbH in Graz 
and partially funded within the COMET K2 Competence Centers for 
Excellent Technologies from the Austrian Federal Ministry for Climate 
Action (BMK), the Austrian Federal Ministry for Digital and Economic 
Affairs (BMDW), the Province of Styria (Dept. 12) and the Styrian 
Business Promotion Agency (SFG). The Austrian Research Promotion 
Agency (FFG) has been authorised for the programme management.

	\bibliography{main}{}
	\bibliographystyle{plain}
		
\end{document}